\documentclass[5p,twocolumn]{elsarticle}

\makeatletter
\def\ps@pprintTitle{%
  \let\@oddhead\@empty
  \let\@evenhead\@empty
  \def\@oddfoot{\reset@font\hfil\thepage\hfil}
  \let\@evenfoot\@oddfoot
}
\makeatother

\usepackage{hyperref}
\usepackage[labelfont=bf,font=normalsize]{caption}
\usepackage{array}
\usepackage{arydshln}
\usepackage{subcaption}
\usepackage{multirow}

\renewcommand{\arraystretch}{1.3} 

\newcommand{\specialcell}[2][c]{%
  \begin{tabular}[#1]{@{}l@{}}#2\end{tabular}}

\usepackage{titlesec}
\titleformat{\paragraph}
  {\normalfont\normalsize\itshape}
  {\theparagraph}{0.5em}{}  

\titlespacing*{\paragraph}
  {0pt}    
  {2ex}    
  {0pt}    

\begin{document}

\begin{frontmatter}
    \title{Federated Learning for MRI-based BrainAGE: a multicenter study on post-stroke functional outcome prediction}
    \author[1]{Vincent Roca\corref{cor1}}
    \author[2]{Marc Tommasi}
    \author[2]{Paul Andrey}
    \author[3]{Aurélien Bellet}
    \author[4]{Markus D. Schirmer}
    \author[5]{Hilde Henon}
    \author[5,6]{Laurent Puy}
    \author[7]{Julien Ramon}
    \author[1,6,7]{Grégory Kuchcinski}
    \author[1,6,7]{Martin Bretzner}
    \author[1,6,8]{Renaud Lopes}
    \author{on behalf of the ARIANES study group\fnref{fnArianes}}

    \cortext[cor1]{Corresponding author at: Lille, 59037 France. E-mail address: vincentroca9@outlook.fr}
    \address[1]{Univ. Lille, CNRS, Inserm, CHU Lille, Institut Pasteur de Lille, US 41 - UAR 2014 - PLBS, F-59000 Lille, France}
    \address[2]{Univ. Lille, Inria, F-59000 Lille, France}
    \address[3]{Univ. Montpellier, Inria, F-34090 Montpellier, France}
    \address[4]{Department of Neurology, Massachusetts General Hospital, Harvard Medical School, Boston, USA}
    \address[5]{CHU Lille, Département de Neurologie vasculaire, F-59037 Lille, France}
    \address[6]{Univ. Lille, Inserm, CHU Lille, U1172 - Lille Neurosciences \& Cognition, F-59037 Lille, France}
    \address[7]{CHU Lille, Département de Neuroradiologie, F-59037 Lille, France}
    \address[8]{CHU Lille, Département de Médecine Nucléaire, F-59037 Lille, France}

    \fntext[fnArianes]{Members of the ARIANES study group: Jean-Pierre Pruvo, Mathieu Masy, Denis Berteloot, Cyril Bruge, Sebastien Verclytte, Emmanuel Michelin, Thierry Stekelorom, and Wael Yacoub.}

    \begin{abstract}
    
        \textbf{Objective:} Brain-predicted age difference (BrainAGE) is a neuroimaging biomarker reflecting brain health, with potential implications for post-stroke recovery. However, training robust BrainAGE models requires large, diverse datasets, often restricted by privacy and regulatory concerns. This study evaluates the performance of federated learning (FL) for BrainAGE estimation in ischemic stroke patients treated with mechanical thrombectomy, and investigates its association with clinical phenotypes and functional outcome.
        
        \textbf{Methods:} We used pre-treatment FLAIR brain images from 1674 stroke patients across 16 hospital centers. We implemented standard machine learning and deep learning models for BrainAGE estimates under three data management strategies: centralized learning (pooled data), FL (local training at each site), and single-site learning. We reported prediction errors and examined associations between BrainAGE and vascular risk factors (e.g., diabetes mellitus, hypertension, smoking), as well as functional outcome at three months post-stroke. Logistic regression evaluated BrainAGE’s predictive value for this outcome, adjusting for age, sex, vascular risk factors, stroke severity, time between MRI and arterial puncture, prior intravenous thrombolysis, and recanalisation outcome.
        
        \textbf{Results:} While centralized learning yielded the most accurate predictions, FL consistently outperformed single-site models. BrainAGE was significantly higher in patients with diabetes mellitus across all models. Comparisons between patients with good and poor functional outcome, and multivariate predictions of these outcome showed the significance of the association between BrainAGE and post-stroke recovery.
        
        \textbf{Conclusion:} FL enables accurate age predictions without data centralization. The strong association between BrainAGE, vascular risk factors, and post-stroke recovery highlights its potential for personalized prognostic modeling in stroke care.
    \end{abstract}

    
\end{frontmatter}

\section{Introduction}

Stroke is a leading cause of mortality and long-term disability. Ischemic strokes account for approximately 87\% of all cases \citep{Capirossi2023}, making them the primary focus of therapeutic advancements. Advances in acute ischemic stroke treatment, such as thrombolysis and mechanical thrombectomy (MT), have significantly improved survival and functional outcome rates. However, these therapies alone do not ensure complete recovery, and many patients require long-term rehabilitation to regain functional abilities. Identifying patients who might need dedicated care is the first step towards personalized medicine and underscores the importance of accurately assessing and predicting post-stroke functional outcome to guide patient management.

Several factors influence post-stroke outcome, including severity and location of stroke, pre-existing comorbidities, and timely access to specialized care. Among these, age is a critical determinant, as older patients often experience worse recovery trajectories due to reduced neuroplasticity and higher rates of comorbid conditions \citep{Drozdowska2019}. However, the effects of aging are highly dependent on past medical history, lifestyle, and genetics and are variable among the subjects. Some will experience healthy, disease-free aging, whereas others will suffer more from the consequences of age-related diseases such as ischemic stroke \citep{Horvath2018}.

In recent years, brain age gap estimation (BrainAGE) has emerged as a biomarker of  brain health and resilience, offering insights into recovery potential that chronological age alone cannot predict \citep{Franke2019}. To compute BrainAGE, machine learning (ML) algorithms are trained to predict an individual's chronological age from neuroimaging data; BrainAGE is then derived from the relative difference between inferred age and chronological age. A lower BrainAGE may reflect greater resilience to functional disabilities and disease severity, while a higher BrainAGE could indicate reduced resilience and an elevated risk of post-stroke disability---thus linking the metric to brain health and resilience.

Traditionally, BrainAGE has been trained on T1-weighted (T1w) MRI scans to assess the risk of aging-related diseases, such as Alzheimer’s disease \citep{Gautherot2021}. Its application to stroke care remains underexplored, despite its potential to offer an innovative approach for predicting patient outcome and guiding personalized treatment strategies. A lower BrainAGE has been associated with a reduced risk of post-stroke cognitive impairment up to 36 months after stroke, even in patients without previous cognitive impairments \citep{Aamodt2023}. \citet{Bretzner2023} instead used fluid-attenuated inversion recovery (FLAIR) sequences, which are more systematically acquired in the acute phase than T1w images, to train a BrainAGE algorithm. They showed that higher BrainAGE derived from FLAIR images was associated with an unfavorable vascular profile and worse post-stroke prognosis.

To fully harness the potential of BrainAGE for post-stroke outcome prediction, it is essential to train ML algorithms on large and diverse datasets. Diversity primarily refers to age range, but also encompasses site-related variability, which influences age prediction accuracy \citep{Bashyam2022,Liu2023}. Accordingly, \citet{Bretzner2023} trained their model by centralizing a large multicenter, international cohort of 4500 patients. However, privacy concerns and regulatory constraints often limit such data sharing across institutions, making federated learning (FL) an innovative and viable solution to address these challenges \citep{Rieke2020}. Unlike traditional centralized learning, where data from multiple centers is pooled into a single repository, FL allows models to be trained locally at each site, sharing only model updates rather than raw data. This decentralized paradigm is particularly advantageous for neuroimaging research \citep{Sandhu2023}.

Previous studies have demonstrated that ML models can achieve comparable performance in age prediction under centralized and federated settings \citep{Basodi2021,Cheshmi2023,Souza2022,Stripelis2021a,Stripelis2021b,Stripelis2022,Stripelis2024}. However, these applications have been restricted to healthy populations, preventing the evaluation of BrainAGE as a biomarker for brain diseases. Moreover, these prior studies did not leverage truly multicentric datasets but rather simulated data distributions across centers, bypassing site-specific variability, which can significantly affect age prediction performance \citep{Roca2023}.

Given these limitations, our study aims to bridge the gap between BrainAGE applied in multicenter stroke cohorts and federated learning-based BrainAGE approaches that address practical barriers to multicenter model development. More specifically, we aim to compare centralized and federated approaches for age prediction in a real-world clinical population and to evaluate the clinical informativeness of BrainAGE obtained under these different training paradigms.

To this end, we implemented BrainAGE estimation using centralized, federated, and single-site training strategies. To account for varying computational capacities across centers, we tested these three approaches using base ML models of different complexities, simulating scenarios ranging from limited to extensive computational capabilities. Leveraging a multicenter dataset of FLAIR images from 16 hospital centers acquired prior to any treatment, we systematically assessed age prediction accuracy, explored associations between BrainAGE and vascular phenotypes, and evaluated its predictive value for post-stroke functional outcome.

\section{Materials and Methods}

\subsection{Ethics}
The ethical committee (Comité de protection des personnes Nord-Ouest IV) classified the study as observational on March 9, 2010, and the committee protecting personal information of the patient approved the study by December 21, 2010 (n° 10.677). The DPO department of the Lille University Hospital attests to the declaration of the implementation methods of this project, in accordance with the applicable personal data protection regulation, in particular the General Data Protection Regulation (UE) 2016/679.

\subsection{Participants and clinical variables}
The cohort collection began in December 2010. However, randomized controlled trials demonstrated the efficacy of MT in 2015. Our imaging cohort therefore focused on patients who benefited from the new standard of care for large vessel occlusion ischemic stroke. Consequently, patients admitted before and after 2015 were not directly comparable. 

Accordingly, we included all consecutive patients admitted to our comprehensive stroke center for MT due to acute ischemic stroke between January 1, 2015, and December 31, 2021. All patients were treated following the latest international guidelines \citep{Powers2018}. FLAIR brain imaging acquired prior to any treatment was obtained for each participant in one of 16 stroke centers in our stroke network in the Hauts-de-France region. Clinical data were prospectively collected by trained vascular neurologists, including demographic characteristics (age, sex assigned at birth) and major vascular risk factors: hypertension (HTN), diabetes mellitus (DM), atrial fibrillation (AF), smoking status (SMK), and hypercholesterolemia (HCL). Stroke severity upon admission was assessed using the National Institutes of Health Stroke Scale (NIHSS). Additional variables included the time interval in minutes between the acquisition of brain MRI and the arterial puncture (P2P), intravenous thrombolysis (IVT), and successful recanalisation (RECA, defined as mTICI $\geq$ 2b on final angiography). Functional outcome were assessed using the modified Rankin Scale (mRS) at 3 months post-stroke.

All FLAIR images were visually inspected, and those with major artifacts (e.g., folding, motion, metal, or severe bias-field inhomogeneity) were excluded. Participants with missing clinical data and those with basilar artery occlusion---due to their distinct and typically poor prognosis \citep{Bagheri2024}---were also excluded.

\subsection{Age prediction}
\subsubsection{Centralized, federated and single-site models}
\label{meth:trainingConfigs}
We studied different ML algorithms for age estimation (Section \ref{meth:ml_approaches}), each implemented under three distinct training configurations:
\begin{itemize}
    \item centralized: The images from all hospital centers were aggregated onto a single machine for model training.
    \item federated: The images were kept separate in each center, centralizing only the training gradients after each local epoch (Section \ref{meth:fl_implementation}).
    \item single-site: Only the images from Center 1---the most represented center in the dataset (Section \ref{res:population})---were used for training.
\end{itemize}

We followed different strategies to obtain age prediction for our \textbf{multicenter evaluation cohort}---consisting of all the images acquired outside Center 1---with each implemented model. For the single-site models, training was performed solely on the Center 1 images, and inference was applied on the remaining images. For the centralized and federated models, we employed five-fold cross-validation on the multicenter evaluation cohort, preserving the distribution of centers across folds. In each fold, the training set included Center 1 images and the training split of the fold, while inference was performed on the remaining patients of the multicenter evaluation cohort.

Comparing the centralized and federated approaches allowed for an evaluation of the replicability of results obtained in a centralized setting---which is supposed to be optimal in terms of prediction performances---with FL. The single-site approach served as an ablated version to assess the added value of multicenter training data.

\subsubsection{Implementation of federated learning}
\label{meth:fl_implementation}
We implemented the FedAvg algorithm \citep{Mcmahan2017} for FL. In FedAvg, at each communication round $t$, a central server sends the current global model parameters $w_t$ to a subset of $K$ participating clients. Each client $k$ then trains a local copy of the model on its private dataset for several local epochs using stochastic gradient descent (SGD), producing updated parameters $w_{t+1}^k$. After local training, the clients transmit their updated model weights back to the server. The server then aggregates these updates using a weighted average: 

$$w_{t+1} = \sum_{k=1}^{K} \alpha_k w_{t+1}^k$$

where $\alpha_k$ is proportional to each client's dataset size (i.e., the number of MR images at each center).

Given the small number of clients (i.e., hospital centers) in our dataset (Section \ref{res:population}), we included all clients in every round, with each one communicating with the server after a single local epoch, as communication costs were a minor concern compared to convergence. 

We used Declearn\footnote{\url{https://gitlab.inria.fr/magnet/declearn/declearn2/-/tree/v2.6.0?ref_type=tags} accessed 2024-09-15}, a Python package designed to facilitate FL by coordinating distributed models through a central server, to implement FedAvg.

\subsubsection{Machine learning approaches}
\label{meth:ml_approaches}
Using the previously defined training configurations (Section \ref{meth:trainingConfigs}), we implemented different regression models for age prediction in order to compare the approaches at different levels of complexity.

\paragraph{Volume-based approach}\label{meth:volBase_ml}\hfill

We applied SynthSeg \citep{Billot2023} to our FLAIR images to extract 32 brain volumes for each image, which we then normalized with the estimated total intracranial volume. These volumes served as input features for linear regression trained using SGD, optimizing the mean absolute error with an L2 regularization term. A min-max scaler was included as a preprocessing step in the model.

We implemented two versions of this model. The first, \textbf{VolSimpleLR}, directly used the 32 volumes as input features. To increase model complexity and capture potential nonlinear patterns, we developed \textbf{VolAugmentedLR}, which expanded the input space by generating all polynomial combinations of the original features up to degree 2 (after min-max scaling). This included each original feature, its square, and all pairwise products, resulting in a total of 560 features.

For the centralized and single-site settings, models were trained for 1000 epochs using inverse scaling learning rate decay, starting at 0.5 for \textit{VolSimpleLR} and 0.07 for \textit{VolAugmentedLR}. In the federated configuration, training proceeded for 1000 rounds with linear learning rate decay (found empirically to be more stable than inverse scaling in this setting), ranging from 0.1 to 0.01 for \textit{VolSimpleLR} and from 0.02 to 0.002 for \textit{VolAugmentedLR}. To accelerate convergence, the intercept was initialized to the mean age in the training set for the centralized model, and to the mean age of Center 1 for the federated and single-site models. For the federated model, the global training-set mean was not used in order to preserve the privacy of age data across centers. 

A five-fold cross-validation procedure was used in the centralized and single-site models to tune the L2 penalty weight by minimizing mean absolute error. The FL models adopted the L2 penalty weights identified in the single-site training.

\paragraph{Radiomic-based approach}\label{meth:radiomicsBase_ml}\hfill

We set up a more complex model, \textbf{RadiomicsLR}, based on a feature set consisting of white-matter (WM) radiomics. To minimize technical variability, we first preprocessed the FLAIR images as follows: (i) skull-stripping with HD-BET \citep{Isensee2019}, (ii) bias correction with N4ITK \citep{Tustison2010}, (iii) linear registration with FSL-FLIRT \citep{Jenkinson2002} (six degrees of freedom), and (iv) z-score normalization of intensities based on the mean and standard deviations computed within the registered brain mask. For registration, we selected a reference image with 0.6 x 0.6 mm of pixel resolution in axial plane and 5.5 mm of slice thickness---an average resolution within our dataset (Section \ref{res:population})---since spatial resampling near the original voxel size enhances the reproducibility of radiomics \citep{Wichtmann2022}.

Next, we extracted WM masks from the registered SynthSeg segmentations and computed 1381 WM radiomics for each image using Pyradiomics \citep{vanGriethuysen2017}. The extraction parameters were based on those used by \citet{Bretzner2023} (availables online, see Section \ref{meth:dataCode}).

The radiomics served as inputs for linear models configured as in Section \ref{meth:volBase_ml}, with reduced learning rates: inverse scaling starting at 0.004 for the centralized and single-site models, and linear decay from 0.01 to 0.001 for the federated model.

\paragraph{Voxel-based approach}\hfill

Finally, we implemented a deep learning model, \textbf{VoxelsCNN}, based on the 3D convolutional neural network architecture described by \citet{Cole2017}. Preprocessing was the same as before radiomics extraction (Section \ref{meth:radiomicsBase_ml}). We modified the max-pooling layers in \citeauthor{Cole2017}'s convolutional blocks to not downsample along the head-foot axis, which is weakly resolved in our images (Section \ref{res:population}). In the federated model, we replaced Batch Normalization with Layer Normalization \citep{Ba2016}, as the latter has been found to perform better in FL \citep{Casella2024}. To accelerate convergence, we initialized the intercept of the output dense layer to the average age in the training set for the centralized model and to the average age in Center 1 for the federated and single-site models.

All models used a batch size of 8. Centralized and single-site models were trained for 1000 epochs using the Adam optimizer \citep{Kingma2014}, with linear learning rate decay from 0.001 to 0.0001. For the federated model, we used plain SGD for the client updates---the strategy employed in FedAvg \citep{Mcmahan2017}, which we found empirically to yield more stable training---using a linear learning rate decay from 0.0005 to 0.00005; we reduced the number of epochs to 500 to limit computational overhead. To save GPU memory and speed up computations, a mixed precision policy was adopted \citep{Micikevicius2017}.

\subsection{Experiments}

\subsubsection{Prediction errors}
We assessed each model’s prediction accuracy by computing the absolute difference (in years) between the actual and predicted ages for each MR image in the multicenter evaluation cohort. Within each of the four ML approaches outlined in Section \ref{meth:ml_approaches}, we compared prediction errors across training configurations (Section \ref{meth:trainingConfigs}) using two-tailed Wilcoxon signed-rank tests, with p-values corrected using the Benjamini–Hochberg (BH) procedure.

\subsubsection{Association between BrainAGE and clinical outcomes}
We derived BrainAGE from the relative difference between actual and predicted age and examined its associations with clinical variables reflecting cardiovascular phenotypes, stroke severity, and post-stroke functional outcome.

\paragraph{BrainAGE computation}\label{meth:brainAge_computation}\hfill

We defined BrainAGE following a standard approach \citep{Beheshti2019}. We first computed the predicted age difference (PAD) by subtracting the actual age from the predicted age for each test image. Next, PAD was linearly regressed on the real age to mitigate regression towards the mean. Finally, we derived BrainAGE by subtracting the regression-estimated PAD from the actual PAD. To approximate real-world application where BrainAGE must be estimated for a single new image, linear regression and PAD correction were performed using ten-fold cross-validation on the multicenter evaluation cohort.

\paragraph{Clinical determinants of accelerated brain aging}\label{meth:clinicalPhenotypes}\hfill

To explore correlations between clinical outcomes and brain aging, we compared BrainAGE across the following binary variables: sex, HTN, DM, AF, SMK, and HCL. Specifically, for each variable, we divided the multicenter evaluation cohort according to its value and tested for statistical differences in BrainAGE using two-tailed Mann–Whitney U tests with BH correction.

\paragraph{Impact of accelerated brain aging on post-stroke functional outcome}\hfill

We used a standard mRS cutoff \citep{Bretzner2023,Nair2025} to dichotomize post-stroke functional outcome into good (mRS at 3 months $\leq$ 2) or poor (mRS at 3 months $>$ 2). Using this categorization, we compared BrainAGE between participants with good and poor outcome using two-tailed Mann-Whitney U tests with BH correction.

Next, we assessed the association between BrainAGE and functional outcome using logistic regression, with dichotomized functional outcome (good vs. poor) as the dependent variable. Independent variables included BrainAGE, age, sex, HTN, DM, AF, SMK, HCL, NIHSS, P2P, IVT, and RECA. Each variable’s contribution was assessed using adjusted odds ratios with corresponding 95\% Wald confidence intervals. Statistical significance was determined using Wald test on logistic regression coefficients. BH correction was applied to the p-values associated to the 12 BrainAGE coefficients.

Furthermore, we performed five-fold cross-validation---stratified by functional outcome---on the multicenter evaluation cohort using logistic regression with the dichotomized functional outcome. Predicted probabilities from the held-out folds were aggregated to compute balanced accuracy (BA) (using a 0.5 threshold) and AUC. 95\% confidence intervals were constructed using bootstrap resampling with 5000 replications.

\subsection{Data and code availability}
\label{meth:dataCode}
The FLAIR images used in this study are not publicly available due to the sensitive nature of the data, which could compromise participant privacy. However, parameters for radiomics extraction and code for training and applying the ML models with the different training configurations are accessible in a public repository: \url{https://github.com/RocaVincent/federated_brain_age}.

\section{Results}

\subsection{Population}
\label{res:population}
\begin{figure*}
    \centering
    \includegraphics[scale=0.4]{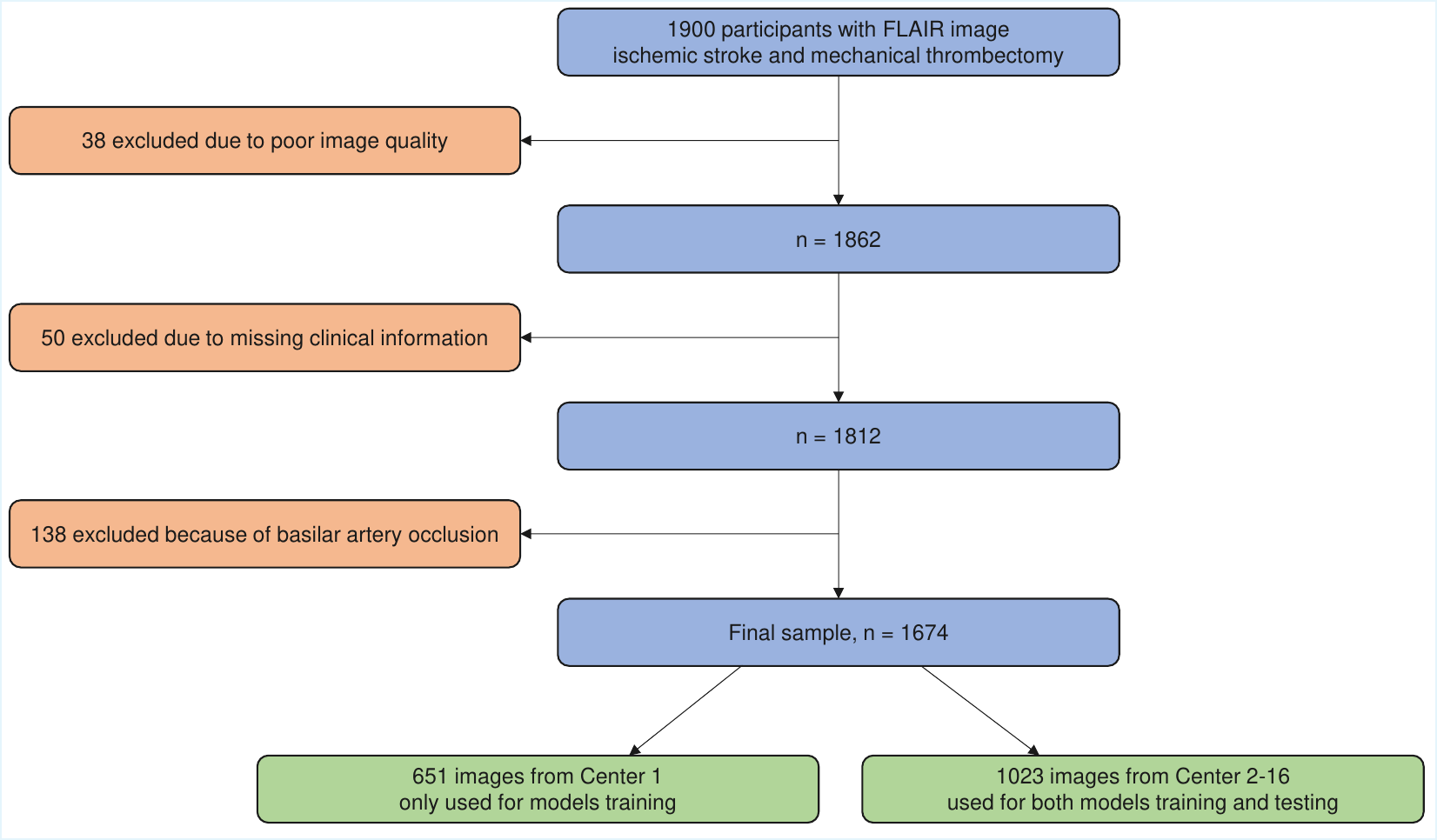}
    \caption{\textbf{Flowchart of data and analysis.}}
    \label{fig:flowchart}
\end{figure*}

1674 participants were included in this study (Figure \ref{fig:flowchart}). Table \ref{tab:center_infos} summarizes the demographic and MRI acquisition characteristics for each of the 16 medical centers. Center 1 contributed the largest portion of the dataset, accounting for 39\% of the cohort. Age distributions across centers were similar, as confirmed by a Kruskal-Wallis H-test on the respective age groups (p = 0.6824). FLAIR images had high in-plane (axial) resolution and typical clinical high slice thickness, with notable heterogeneity observed across centers. Among the clinical variables, P2P and IVT were significantly associated with the acquisition center (\ref{appendix_cliByCenter}).

\begin{table*}
    \centering
    \caption{\textbf{Demographic and MRI acquisition characteristics in the study population.}}
    \begin{tabular}{>{\raggedright}p{0.10\linewidth}>{\raggedright}p{0.128\linewidth}>{\raggedright}p{0.128\linewidth}>{\raggedright}p{0.14\linewidth}>{\raggedright}p{0.10\linewidth}>{\raggedright}p{0.128\linewidth}>{\raggedright\arraybackslash}p{0.128\linewidth}}
        \hline
        \textbf{medical center}&\textbf{nb of participants}&\textbf{age, years}$^\#$&\textbf{manufacturer}$^\dag$&\textbf{field strength, Tesla}$^\dag$&\textbf{in-plane pixel size, mm$^{2\#}$}&\textbf{slice thickness, mm}$^\#$\\
        \hline
        Center 1&651&69.90 ± 15.37&Philips (650); Siemens (1)&1.5 (644); 3.0 (7)&0.89 ± 0.06&5.17 ± 0.44\\
        Center 2&184&70.64 ± 12.78&GE (183); Siemens (1)&1.5 (182); 3.0 (2)&0.50 ± 0.04&5.95 ± 0.48\\
        Center 3&135&69.55 ± 14.99&GE (79); Siemens (56)&1.5 (122); 3.0 (13)&0.59 ± 0.16&5.91 ± 0.21\\
        Center 4&84&73.52 ± 14.02&GE (66); Siemens (18)&3.0 (75); 1.5 (9)&0.58 ± 0.16&5.38 ± 0.40\\
        Center 5&83&70.54 ± 15.52&GE (40); Siemens (43)&3.0 (46); 1.5 (37)&0.45 ± 0.06&5.53 ± 0.28\\
        Center 6&82&79.70 ± 14.10&Siemens (76); GE (6)&1.5 (66); 3.0 (16)&0.51 ± 0.21&5.60 ± 0.46\\
        Center 7&79&70.97 ± 13.36&GE&1.5&0.50 ± 0.00&6.16 ± 0.36\\
        Center 8&66&70.18 ± 16.73&GE&1.5&0.58 ± 0.17&6.00 ± 0.00\\
        Center 9&63&69.37 ± 13.94&Philips (62); \quad GE (1)&1.5&0.62 ± 0.07&5.54 ± 0.18\\
        Center 10&62&68.85 ± 13.50&Siemens&1.5&0.57 ± 0.23&5.78 ± 0.44\\
        Center 11&46&69.22 ± 12.70&Siemens&1.5 (37); 3.0 (9)&0.78 ± 0.04&4.97 ± 0.18\\
        Center 12&44&70.07 ± 15.12&Philips&3.0 (41); 1.5 (3)&0.57 ± 0.13&5.00 ± 0.98\\
        Center 13&32&67.28 ± 12.05&Siemens (17); GE (15)&1.5&0.70 ± 0.21&5.45 ± 0.53\\
        Center 14&30&69.53 ± 16.13&Siemens&1.5 (26); 3.0 (4)&0.79 ± 0.13&4.91 ± 0.71\\
        Center 15&23&64.91 ± 17.76&Siemens (16); Philips (7)&1.5&0.80 ± 0.10&4.93 ± 0.35\\
        Center 16&10&70.70 ± 16.40&Philips&1.5&0.63 ± 0.11&4.93 ± 0.35\\
        \hdashline
        Total&1674&70.03 ± 14.71&Philips (773); GE (535); Siemens (366)&1.5 (1461); 3.0 (213)&0.69 ± 0.20&5.47 ± 0.57\\
        \hline
        \multicolumn{7}{@{}l}{$^\#$ Values are expressed as mean $\pm$ standard deviation.}\\
        \multicolumn{7}{@{}l}{$^\dag$ The number of participants is indicated in brackets if there are several options.}\\
    \end{tabular}
    \label{tab:center_infos}
\end{table*}

Table \ref{tab:cliTestSet} summarizes the demographic and clinical characteristics of the multicenter evaluation cohort participants.

\begin{table*}
    \centering
    \caption{\textbf{Demographic and clinical characteristics in the multicenter evaluation cohort.}}
    \begin{tabular}{ll}
         \hline
         \textbf{sample size,} nb of participants&1023\\
         \textbf{age,} years, mean ± standard deviation&70.10 ± 14.27\\
         \textbf{sex,} \% of male&46\\
         \textbf{hypertension,} \%&68\\
         \textbf{diabetes mellitus,} \%&21\\
         \textbf{atrial fibrillation}, \%&37\\
         \textbf{smoking,} \%&17\\
         \textbf{hypercholesterolemia,} \%&41\\
         \specialcell[c]{\textbf{NIH Stroke Scale upon admission,}\\median (interquartile range)}&16 (8)\\
         \specialcell[c]{\textbf{time interval between MRI and the arterial puncture,}\\minutes, mean ± standard deviation}&148.18 ± 40.27\\
         \textbf{previous intravenous thrombolysis,} \%&62\\
         \textbf{successful recanalisation,} \%&83\\
         \specialcell[c]{\textbf{Modified Rankin Scale at three months,}\\median (interquartile range); \% of good outcome ($\leq$ 2)}&3 (3); 38\\
         \hline
    \end{tabular}
    \label{tab:cliTestSet}
\end{table*}

\subsection{Examples of BrainAGE estimation}
\begin{figure*}
    \centering
    \begin{subfigure}[c]{\linewidth}
        \hspace{2em}
        \begin{minipage}[c]{0.2\linewidth}
            \includegraphics[scale=1]{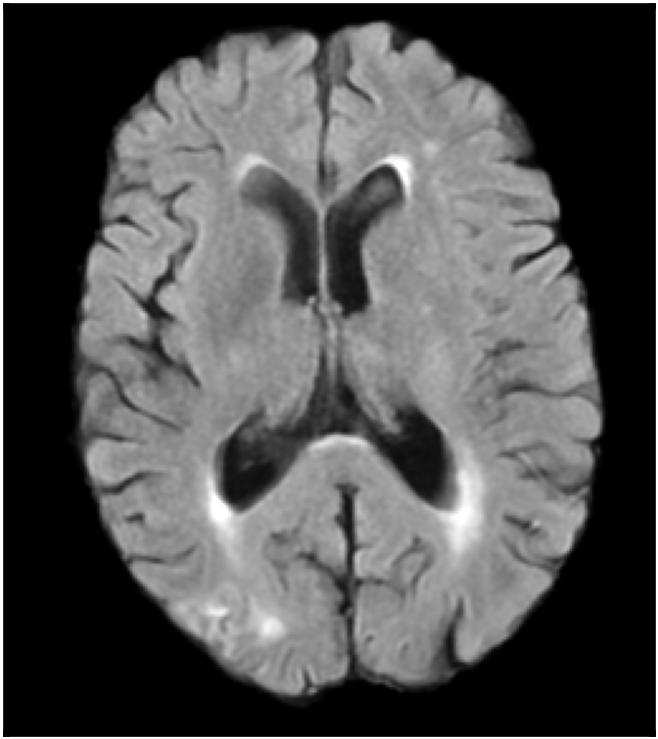}
        \end{minipage}
        \hspace{0.1\linewidth}
        \begin{minipage}[c]{0.5\linewidth}
            \begin{tabular}{llll}
                 &\textbf{centralized}&\textbf{federated}&\textbf{single-site}\\
                 \textbf{VolSimpleLR}&+12.60&+12.54&+13.48\\
                 \textbf{VolAugmentedLR}&+12.72&+14.39&+14.86\\
                 \textbf{RadiomicsLR}&+11.94&+11.98&+9.94\\
                 \textbf{VoxelsCNN}&+9.78&+6.33&+7.53\\
            \end{tabular}
        \end{minipage}
        \caption{Participant aged 62 with positive BrainAGE for all methods.}
        \label{fig:high_BrainAGE}
    \end{subfigure}
    \vspace{1em}
    
    \begin{subfigure}[c]{\linewidth}
        \hspace{2em}
        \begin{minipage}[c]{0.2\linewidth}
            \includegraphics[scale=1]{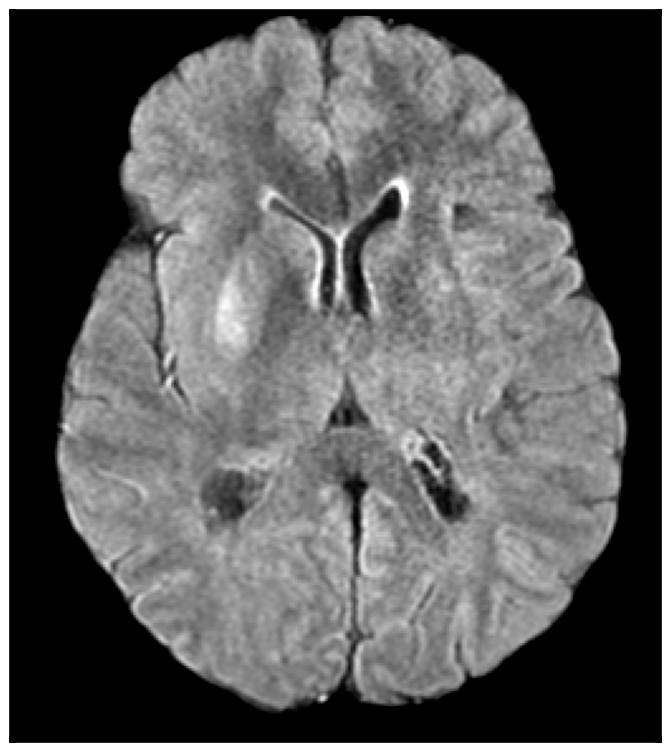}
        \end{minipage}
        \hspace{0.1\linewidth}
        \begin{minipage}[c]{0.5\linewidth}
            \begin{tabular}{llll}
                 &\textbf{centralized}&\textbf{federated}&\textbf{single-site}\\
                 \textbf{VolSimpleLR}&-8.69&-9.39&-9.49\\
                 \textbf{VolAugmentedLR}&-9.69&-9.92&-8.01\\
                 \textbf{RadiomicsLR}&-15.91&-10.61&-9.11\\
                 \textbf{VoxelsCNN}&-1.65&-13.14&-10.90\\
            \end{tabular}
        \end{minipage}
        \caption{Participant aged 68 with negative BrainAGE for all methods.}
        \label{fig:low_BrainAGE}
    \end{subfigure}
    \caption{\textbf{BrainAGE estimated on two FLAIR images of the multicenter evaluation cohort.} The middle axial slice after skull-stripping, inhomogeneity correction and linear registration is shown for each one, along with BrainAGE for each implemented model.}
\end{figure*}

Figure \ref{fig:high_BrainAGE} depicts the brain of a patient admitted for a left middle cerebral artery stroke. The image reveals substantial cortical and subcortical atrophy, with enlarged sulci and ventricles. Confluent periventricular white matter hyperintensities and a prior stroke lesion in the right occipital lobe are also visible. The corresponding BrainAGE was positive across all methods.

On the contrary, Figure \ref{fig:low_BrainAGE} shows the brain of a patient who, notwithstanding a right deep middle artery stroke, exhibits a good brain health with a maintained parenchymal trophicity, collapsed sulci, thin ventricles, and minimal WM hyperintensities. The corresponding BrainAGE was negative for all methods.

\subsection{Prediction errors}
\label{ref:predictionErrors}
\begin{figure*}
    \centering
    \begin{minipage}[c]{0.4\linewidth}
        \includegraphics[scale=1]{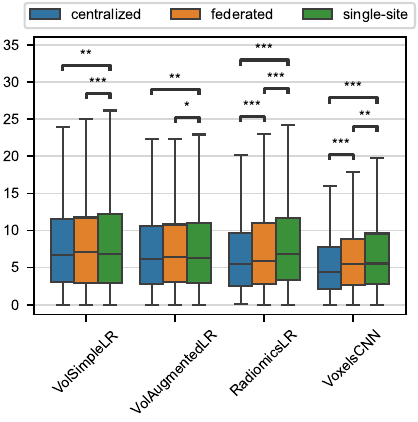}
    \end{minipage}
    \begin{minipage}[c]{0.55\linewidth}
        \footnotesize
        \vspace{-3em}
        \begin{tabular}{llll}
            \vspace{2em}
            &\textbf{centralized}&\textbf{federated}&\textbf{single-site}\\
            \vspace{2em}
            \textbf{VolSimpleLR}&8.03 ± 6.22&8.13 ± 6.43&8.49 ± 6.91\\
            \vspace{2em}
            \textbf{VolAugmentedLR}&7.44 ± 5.87&7.56 ± 5.85&7.65 ± 6.07\\
            \vspace{2em}
            \textbf{RadiomicsLR}&6.85 ± 5.57&7.35 ± 5.97&8.60 ± 8.65\\
            \vspace{2em}
            \textbf{VoxelsCNN}&5.53 ± 4.50&6.46 ± 4.99&7.04 ± 5.94\\
        \end{tabular}
    \end{minipage}
    \caption{\textbf{Absolute age prediction errors in the multicenter evaluation cohort.} Boxplots with significant paired comparisons of errors ($^*$: p $<$ 0.05; $^{**}$: p $<$ 0.01; $^{***}$: p $<$ 0.001) and mean ± standard deviation of the errors are shown.}
    \label{fig:errors_pred}
\end{figure*}

Figure \ref{fig:errors_pred} shows the absolute age prediction errors. The comparison of the four ML approaches reveals that greater model complexity led to more precise age estimations. When comparing the three training methods, the centralized approach achieved the lowest errors, with statistically significant differences except when compared to the federated \textit{VolSimpleLR} and \textit{VolAugmentedLR} models. The federated models exhibited higher errors than the centralized models but significantly outperformed the single-site versions in pairwise comparisons. Notably, \textit{RadiomicsLR} outperformed the two volume-based approaches in both the centralized and federated configurations, but demonstrated more variability and higher errors in the single-site configuration.

\subsection{Clinical phenotype and brain aging}
\label{res:phenotypeComparison}
\begin{figure*}
    \centering
    \includegraphics[scale=1]{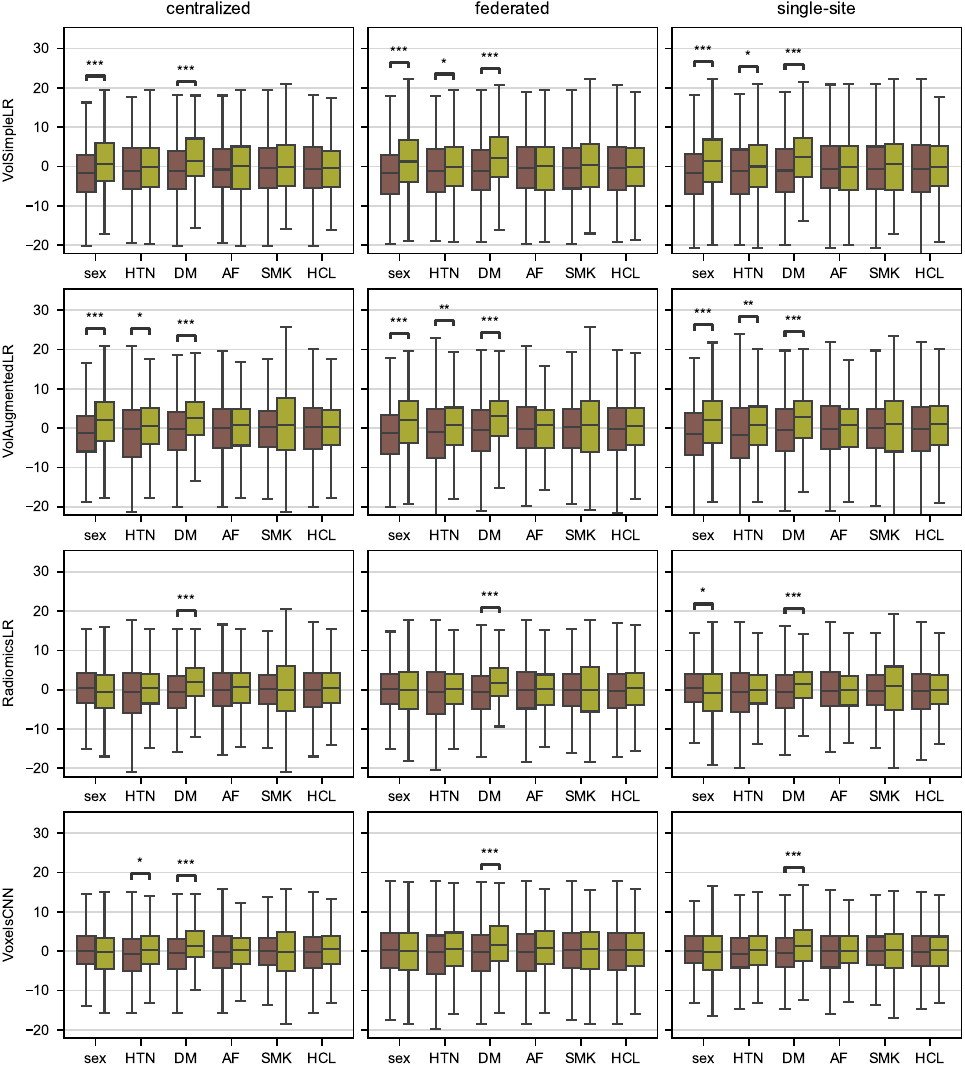}
    \includegraphics[scale=1]{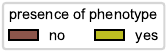}
    \caption{\textbf{Comparison of BrainAGE for each clinical phenotype in the multicenter evaluation cohort.} For the sex variable, the presence or absence of the phenotype corresponds to male and female, respectively. Asterisks indicate significant unpaired comparisons of BrainAGE ($^*$: p $<$ 0.05; $^{**}$: p $<$ 0.01; $^{***}$: p $<$ 0.001).\\
    Abbreviations: HTN: hypertension; DM: diabetes mellitus; AF: atrial fibrillation; SMK: smoking; HCL: hypercholesterolemia.}
    \label{fig:phenotyppe_BrainAGE}
\end{figure*}

Several observations emerge from the BrainAGE comparisons by clinical phenotype (Figure \ref{fig:phenotyppe_BrainAGE}). Patients with DM exhibited significantly higher BrainAGE across all models. For HTN and AF, higher median BrainAGEs were observed across all models, but statistical significance was only reached for HTN with some models. This tendency was also observed with SMK and HCL, but with some exceptions (e.g., lower median BrainAGE obtained for SMK with the centralized \textit{RadiomicsLR} model). Regarding sex, males exhibited significantly higher BrainAGE in the \textit{VolSimpleLR} and \textit{VolAugmentedLR} models. In contrast, \textit{RadiomicsLR} and \textit{VoxelsCNN} models yielded higher median BrainAGEs for females, with statistical significance for the single-site \textit{RadiomicsLR} model.

\subsection{Post-stroke functional outcome and brain aging}
\begin{figure*}
    \centering
    \includegraphics[scale=1]{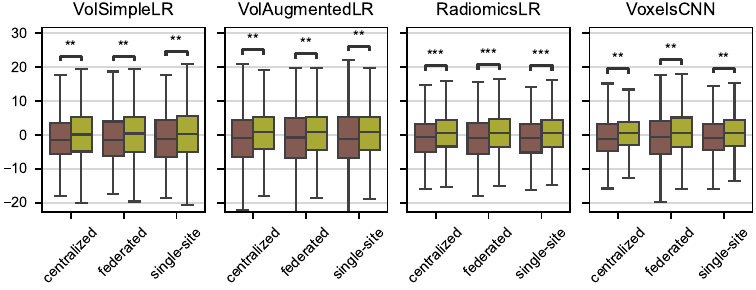}\\
    \includegraphics[scale=1]{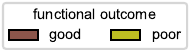}
    \caption{\textbf{Comparison of BrainAGE between good and poor functional outcome in the multicenter evaluation cohort.} Asterisks indicate significant unpaired comparisons of BrainAGE ($^*$: p $<$ 0.05; $^{**}$: p $<$ 0.01; $^{***}$: p $<$ 0.001).}
    \label{fig:mrs_BrainAGE}
\end{figure*}

All implemented models exhibited significantly higher BrainAGE in the patients with poor functional outcome, regardless of whether centralized, federated, or single-site training was used (Figure \ref{fig:mrs_BrainAGE}).

The contributions of BrainAGE in logistic regression of good functional outcome are summarized in Table \ref{tab:BrainAGE_OR}. Across all models, the odds ratio for BrainAGE was below 1, indicating that higher BrainAGE was associated with poorer post-stroke outcome. The corresponding regression coefficients were significantly different from 0. Furthermore, BrainAGE had a lower odds ratio with the \textit{RadiomicsLR} and \textit{VoxelsCNN} models. Similar BA and AUC values were obtained across models, except for the model without BrainAGE, which showed slightly lower performance for functional outcome prediction.

\begin{table*}
    \centering
    \caption{\textbf{BrainAGE odds ratio, balanced accuracy and AUC in multiple logistic regression of good functional outcome in the multicenter evaluation cohort.}}
    \begin{tabular}{>{\raggedright}p{0.17\linewidth}>{\raggedright}p{0.08\linewidth}>{\raggedright}p{0.05\linewidth}>{\raggedright}p{0.14\linewidth}>{\raggedright}p{0.05\linewidth}>{\raggedright}p{0.14\linewidth}>{\raggedright}p{0.05\linewidth}>{\raggedright\arraybackslash}p{0.14\linewidth}}
        \hline
        &&\multicolumn{2}{l}{\textbf{centralized}}&\multicolumn{2}{l}{\textbf{federated}}&\multicolumn{2}{l}{\textbf{single-site}}\\
        \hdashline
        \multirow{3}{*}{\textbf{VolSimpleLR}}&OR&0.973$^{*}$&$[0.954,0.993]$&0.978$^*$&$[0.960,0.996]$&0.981$^*$&$[0.963,0.998]$\\
        &BA&0.676&$[0.646,0.703]$&0.672&$[0.642,0.700]$&0.672&$[0.642,0.700]$\\
        &AUC&0.773&$[0.743,0.801]$&0.772&$[0.743,0.800]$&0.772&$[0.742,0.800]$\\
        \hdashline
        \multirow{3}{*}{\textbf{VolAugmentedLR}}&OR&0.973$^{*}$&$[0.954,0.992]$&0.975$^*$&$[0.956,0.994]$&0.973$^{*}$&$[0.955,0.992]$\\
        &BA&0.679&$[0.649,0.706]$&0.669&$[0.639,0.698]$&0.676&$[0.646,0.704]$\\
        &AUC&0.773&$[0.743,0.801]$&0.773&$[0.743,0.801]$&0.774&$[0.744,0.802]$\\
        \hdashline
        \multirow{3}{*}{\textbf{RadiomicsLR}}&OR&0.962$^{**}$&$[0.940,0.984]$&0.964$^{**}$&$[0.943,0.986]$&0.966$^{**}$&$[0.944,0.988]$\\
        &BA&0.674&$[0.644,0.702]$&0.677&$[0.648,0.706]$&0.679&$[0.650,0.707]$\\
        &AUC&0.777&$[0.748,0.804]$&0.776&$[0.746,0.803]$&0.774&$[0.745,0.802]$\\
        \hdashline
        \multirow{3}{*}{\textbf{VoxelsCNN}}&OR&0.960$^{**}$&$[0.936,0.984]$&0.972$^{*}$&$[0.952,0.993]$&0.955$^{**}$&$[0.931,0.980]$\\
        &BA&0.673&$[0.644,0.701]$&0.675&$[0.646,0.703]$&0.672&$[0.643,0.701]$\\
        &AUC&0.775&$[0.746,0.803]$&0.773&$[0.744,0.801]$&0.778&$[0.748,0.805]$\\
        \hline
        \multicolumn{8}{p{0.95\linewidth}}{Odds ratio (OR) of BrainAGE with the corresponding 95\% confidence interval is reported. Asterisks indicate that the BrainAGE regression coefficient was significantly different from 0 ($^*$: p $<$ 0.05; $^{**}$: p $<$ 0.01; $^{***}$: p $<$ 0.001)}\\
        \multicolumn{8}{p{0.95\linewidth}}{Age, sex, hypertension, diabetes mellitus, atrial fibrillation, smoking, hypercholesterolemia, NIH Stroke Scale upon admission, time interval in minutes between MRI and the arterial puncture, previous intravenous thrombolysis, and successful recanalisation were the other independent variables.}\\
        \multicolumn{8}{p{0.95\linewidth}}{Balanced accuracy (BA) and AUC were computed using five-fold cross-validation on the multicenter evaluation cohort. The corresponding 95\% confidence intervals are reported. The values without BrainAGE as an independent variable were $BA = 0.663 \, [0.634,0.691]$, $AUC = 0.770 \, [0.741,0.799]$.}\\
    \end{tabular}
    \label{tab:BrainAGE_OR}
\end{table*}

\ref{appendix_oddsRatio} presents the contributions of the other independent variables. Age, NIHSS, P2P, IVT and RECA were consistently significant predictors of the functional outcome. Sex was also significant with certain models.

\section{Discussion}
By leveraging a real-world multicenter ischemic stroke imaging cohort, this study investigated the feasibility and clinical relevance of FL for training FLAIR-based BrainAGE models across heterogeneous clinical sites. While BrainAGE has previously been explored as a biomarker of brain health and post-stroke outcomes, and FL has demonstrated promising performance for medical imaging applications, the combination of both approaches in a clinical multicenter stroke population has remained largely unexplored. Our results showed that, although centralized models achieved slightly higher age prediction accuracy, FL substantially outperformed single-site training, demonstrating its ability to exploit multicenter data variability without requiring data centralization. Furthermore, analyses of BrainAGE in relation to clinical phenotypes and its role as a predictor of post-stroke functional outcome revealed comparable findings between the different implemented methods.

\subsection{Age prediction performance}
For the four implemented ML models, lower prediction errors were observed with centralized learning, as expected. The absence of statistical significance in the comparison with FL for the two volume-based ML approaches is likely attributable to the model's simplicity (32 input volumes). Thus, centralized learning remains the upper-bound reference when data pooling is feasible, and FL is not intended to fully match this performance. Rather, its value lies in achieving sufficient reliability to preserve clinically relevant associations between BrainAGE, vascular risk factors, and functional outcomes, while avoiding data transfer between hospitals.

Although centralized models performed best, FL consistently outperformed single-site learning, underscoring its value when data centralization is not feasible. FL performances could potentially improve with techniques designed to address inter-client heterogeneity \citep{Li2020,Karimireddy2020,Wang2020}. However, \citet{Li2022} found that these methods outperform FedAvg only in some cases, particularly when label distributions vary across centers, whereas age distributions were similar in our study. Furthermore, we conducted additional experiments that supported this observation, showing that prediction errors were not reduced when using FedProx \citep{Li2020} instead of FedAvg (\ref{appendix_fedprox}).

Prediction errors also indicate that more complex models tend to provide more precise predictions, which is consistent with previous findings \citep{Wu2024}. However, under single-site learning, \textit{RadiomicsLR} performed worse than \textit{VolSimpleLR} and \textit{VolAugmentedLR}, despite its greater complexity (1381 vs. 32 input features). This can be attributed to the low inter-scanner reproducibility of radiomics computed on FLAIR brain images \citep{Pandey2020}, emphasizing the advantages of multicenter training datasets.

\subsection{Clinical training data}
\label{disc:cliTraining}
Even when BrainAGE is applied to pathological subjects, the widely adopted practice is to train predictive models on a healthy population \citep{Gautherot2021,Aamodt2023,Bashyam2020,Biondo2022,Richard2020}. As an exception, \citet{Bretzner2023} trained their model with MR images from stroke patients. In addition to the demonstrated clinical relevance, the authors suggested that their BrainAGE could better translate to routine clinical care. We adopted the same strategy because, rather than quantifying deviation from normative healthy aging, we aimed to investigate whether inter-individual variation in brain health within a stroke population is associated with vascular risk factors and post-stroke functional outcome. Consequently, the biological interpretation of BrainAGE also changes: positive BrainAGE reflects brains that appear older than expected for stroke patients, rather than relative to a healthy reference population.

Moreover, using a training set representative of the intended application population may reduce unexpected inference behaviors, such as confounding by stroke lesions. Consistent with this hypothesis, we found no significant association between lesion volume and BrainAGE (\ref{appendix_lesionVolsVSbrainAGE}), suggesting that lesion burden is unlikely to be the primary determinant of the estimated BrainAGE, which instead reflects age-related brain characteristics.

Additional comparisons with a state-of-the-art BrainAGE model trained exclusively on healthy participants are presented in \ref{appendix_sota_comparison}. While both models achieved similar age prediction accuracy, our clinically trained model showed a stronger association between BrainAGE and functional outcome and a weaker association with lesion volume, further supporting the potential value of training on a representative clinical population.

\subsection{BrainAGE and clinical phenotypes}
The comparison of BrainAGE between patients with and without each clinical phenotype revealed a significant difference for DM, regardless of the BrainAGE model used. This aligns with the findings of \citet{Bretzner2023}. However, unlike \citeauthor{Bretzner2023}, we did not observe significant BrainAGE differences for HTN, AF, or SMK, except with certain models (e.g., HTN with the \textit{VolAugmentedLR} models). The discrepancy in results could be attributed to the difference in sample sizes (n = 1023 vs. 4163), which affects statistical power. Another explanation is the age disparity observed for the non-significant phenotypes (including HCL), whereas age distributions were similar between DM and non-DM patients (\ref{appendix_agePhenotype}). Indeed, through the subtraction of chronological age from the predicted age and subsequent PAD correction (Section \ref{meth:brainAge_computation}), BrainAGE is designed to be independent of chronological age. As a result, if a biological difference is strongly associated with an age difference, it is unlikely to correspond with a BrainAGE difference. Besides, among the clinical phenotypes included, DM was the most predictive of post-stroke functional outcome in our multivariate analyses (\ref{appendix_oddsRatio}). This supports the hypothesis that biological differences unrelated to age were only discriminant between DM and non-DM patients, explaining why BrainAGE comparisons were significant only for this phenotype. 

The significant BrainAGE differences we observed between male and female patients with certain models did not align with previous literature \citep{Bretzner2023,Sanford2022}. However, we did not investigate the causes of these differences as (i) there were significant age differences between males and females in our dataset (\ref{appendix_agePhenotype}), making relevant comparisons difficult, and (ii) the direction of the difference mainly depended of the ML model and not the training configuration (higher BrainAGE for males with the \textit{VolSimpleLR} and \textit{VolAugmentedLR} models vs. higher BrainAGE for females with the \textit{RadiomicsLR} and \textit{VoxelsCNN} models).

Differences between volume-based (\textit{VolSimpleLR}, \textit{VolAugmentedLR}) and radiomics- and voxel-based (\textit{RadiomicsLR}, \textit{VoxelsCNN}) approaches were also observed for HTN: volume-based models generally showed higher BrainAGE in HTN patients, whereas associations were less consistent for radiomics- and voxel-based models. These approaches may capture lesion burden, WM disease, atrophy, or vascular injury differently. Therefore, model-dependent associations, such as those between BrainAGE and sex or HTN, should be interpreted with caution. Consistent with this, \citet{Dular2024} reported associations between clinical variables and BrainAGE that were not consistently significant across models and were therefore considered potential false positives.

To mitigate the impact of age imbalance, we repeated the analyses using age-matched subgroups. The results were consistent with those obtained on the full multicenter evaluation cohort (\ref{appendix_ageMatchComparisons}).

\subsection{BrainAGE and post-stroke functional outcome}
Patients with good functional outcome exhibited significantly lower BrainAGE across all models. BrainAGE also remained significantly associated with functional outcome in logistic regression analyses after adjusting for vascular risk factors and post-stroke treatment variables. Notably, BrainAGE estimates derived from the \textit{RadiomicsLR} and \textit{VoxelsCNN} models showed stronger associations with functional outcome than those from the \textit{VolSimpleLR} and \textit{VolAugmentedLR} models. Examining age prediction errors alongside these results suggests that more accurate age predictions yield BrainAGE measures with greater clinical relevance. This observation is consistent with prior work reporting stronger associations between BrainAGE and disease severity for more accurate predictive models \citep{Lee2025}, as well as highly accurate models (MAE $<$ 1 year) successfully applied to dementia detection \citep{Cap2025}.

However, this relationship is not straightforward. Our single-site \textit{RadiomicsLR} and \textit{VoxelsCNN} models, despite higher prediction errors, still produced BrainAGE values that were strongly associated with functional outcome. Furthermore, multivariable logistic regression models incorporating different BrainAGE variants showed similar predictive performance. Together, these findings suggest that age prediction accuracy is neither sufficient nor necessary for clinical relevance. Previous studies have similarly reported that the most accurate age prediction models do not always yield the most clinically informative BrainAGE estimates \citep{Bashyam2020,Dular2024}. Moreover, the improvement in predictive performance obtained by including BrainAGE---as compared with models without BrainAGE---was modest and should therefore be interpreted as evidence that BrainAGE provides complementary prognostic information rather than a clinically transformative gain in discrimination. More broadly, even when BrainAGE is derived from highly accurate age prediction models trained on large multicenter datasets, its predictive value for health outcomes may remain limited \citep{Tan2025}.

While improvements in age prediction accuracy may not systematically translate into greater clinical utility, large and heterogeneous datasets remain essential to limit overfitting in machine learning, particularly for deep models \citep{Bento2022}. In this context, FL provides an important opportunity to leverage multicenter data to improve model generalizability and reduce site-specific biases \citep{Sandhu2023}.

\subsection{Limitations and perspectives}
We compared models trained using centralized and federated learning with those trained on a single center to assess the benefit of aggregating data from multiple sites. This comparison could be criticized, as single-site models may be affected by site-specific biases, whereas centralized and federated models are exposed to data from all centers during training. A leave-one-center-out (LOCO) strategy would have been a suitable way to mitigate this limitation, but it would have required an extensive number of trainings—given the 16 centers—which was computationally unfeasible for the FL-based deep learning models. Nevertheless, as described in \ref{appendix_trainingConfigs}, we performed supplementary analyses on the linear models, including LOCO for the centralized and federated configurations, within-site cross-validation for the single-site configuration, and a comparison between single-site models trained on Center 1 and those trained on other centers. Across all settings, the centralized and federated approaches consistently outperformed the single-site models, and the single-site models trained on Center 1 outperformed those trained on other centers.

Although the multicenter models---whether centralized or federated---demonstrated better generalization performance than the single-site models under a LOCO configuration (\ref{appendix_trainingConfigs}), their ability to generalize to arbitrary stroke cohorts is not guaranteed. Indeed, all data used in this study originated from the same regional cohort, potentially underestimating the extent of dataset shift. Future work should evaluate these models on more diverse external datasets to better assess their robustness to such shifts.

Another potential criticism is that the centralized and federated models were trained on larger datasets than their single-site counterparts. However, as \citet{Stripelis2024} emphasize, the purpose of federated learning is precisely to leverage distributed data across multiple centers. Therefore, comparing federated models with centralized models trained on smaller datasets remains appropriate, as it reflects the practical advantage of FL in multi-center clinical research.

In Section \ref{disc:cliTraining}, we emphasized the potential benefits of using a clinical dataset for BrainAGE model training. Nonetheless, the standard practice of training and evaluating age prediction models on healthy subjects may produce more accurate estimates and allow for more reliable comparisons across training configurations, given the typically lower and less variable prediction errors.

Additionally, using BrainAGE algorithms in stroke patients involves considering stroke lesions. However, rapid lesion changes are commonly observed in MR images during the acute phase \citep{Vert2017}, and the time interval between symptom onset and imaging completion is often unknown \citep{Thomalla2020}, making it particularly variable in a multicenter dataset. Although we accounted for the time from imaging completion to treatment initiation—which is correlated with post-stroke functional outcome \citep{Sun2013}—, the variable delay between symptom onset and imaging could affect the consistency of BrainAGE estimations. Moreover, even though we found no significant correlation between BrainAGE and stroke lesion volume (\ref{appendix_lesionVolsVSbrainAGE}), brain lesions can hinder automatic tools used to preprocess MR images \citep{Bey2024}. Future studies could explore segmentation tools to mask out lesions, allowing BrainAGE to be estimated independently of lesion-related effects.

In our dataset, age distributions were similar between the centers. Future studies may evaluate FL-based BrainAGE on stroke cohorts with greater heterogeneity in age distributions. Such variability would make age prediction more challenging under FL \citep{Stripelis2024}.

\section{Conclusion}
To build on previous studies demonstrating the utility of BrainAGE as a biomarker for brain disorders, we trained BrainAGE models on FLAIR images from a clinical stroke cohort using FL. FL allows the development of machine learning models on multicenter datasets without requiring data centralization, making it a promising solution to address data-sharing regulations. In our experiments, FL consistently outperformed single-site training. Moreover, the observed associations between BrainAGE and vascular risk factors, along with its predictive value for post-stroke functional outcome, indicate that FL-derived BrainAGE may be as clinically informative as those obtained through centralized training. Together, these findings demonstrate that FL can preserve the clinical informativeness of BrainAGE while overcoming important barriers to multicenter model development, supporting its use as a practical framework for distributed neuroimaging research and future clinical AI applications.

\section*{Acknowledgements}
The Lille University Hospital, leading the StrokeAge project, receives financial support from the Hauts-de-France region as part of the call for projects “Recherche clinique dans les établissements de santé en région Hauts-de-France”.

\section*{Declaration of competing interests}
Markus D. Schirmer is supported by the Heitman Stroke Foundation and NIA R21AG083559.

\appendix

\section{Clinical characteristics by center in the study population}
\label{appendix_cliByCenter}
\setcounter{table}{0}
Table \ref{tab:cliByCenter} presents the distribution of clinical variables across centers in the study dataset. To assess the dependence between each variable and hospital centers, we conducted Kruskal-Wallis H tests for continuous variables (NIHSS, P2P and mRS at three months) and Pearson’s chi-squared tests with Yates’s correction for binary variables. BH correction was applied.

\begin{table*}
    \small
    \centering
    \caption{\textbf{Clinical characteristics by center in the study population.}}
    \begin{tabular}{>{\raggedright}p{0.085\linewidth}@{\hspace{6pt}}>{\raggedright}p{0.06\linewidth}@{\hspace{6pt}}>{\raggedright}p{0.06\linewidth}@{\hspace{6pt}}>{\raggedright}p{0.06\linewidth}@{\hspace{6pt}}>{\raggedright}p{0.06\linewidth}@{\hspace{6pt}}>{\raggedright}p{0.06\linewidth}@{\hspace{6pt}}>{\raggedright}p{0.06\linewidth}@{\hspace{6pt}}>{\raggedright}p{0.08\linewidth}@{\hspace{6pt}}>{\raggedright}p{0.08\linewidth}@{\hspace{6pt}}>{\raggedright}p{0.06\linewidth}@{\hspace{6pt}}>{\raggedright}p{0.07\linewidth}@{\hspace{6pt}}>{\raggedright\arraybackslash}p{0.08\linewidth}}
    \hline
    \textbf{medical center}&\textbf{sex, \%}$^\#$&\textbf{HTN, \%}&\textbf{DM, \%}&\textbf{AF, \%}&\textbf{SMK, \%}&\textbf{HCL, \%}&\textbf{NIHSS, median}&\textbf{P2P, median} $^{***}$&\textbf{IVT, \%} $^{***}$&\textbf{RECA, \%}&\textbf{mRS\_m3, median}\\
    Center 1&45&66&21&41&19&42&15&46&45&80&3\\
    Center 2&43&64&29&37&12&46&17&130.5&55&81&3\\
    Center 3&45&67&18&38&19&45&16&134&50&81&3\\
    Center 4&49&79&13&45&13&35&19&110.5&57&83&3\\
    Center 5&47&67&24&40&13&34&16&135&70&87&3\\
    Center 6&48&65&22&40&21&34&15&183&66&89&3\\
    Center 7&44&65&19&37&19&44&16&157&70&86&3\\
    Center 8&42&68&11&35&15&41&16&141&68&73&4\\
    Center 9&57&73&19&30&24&48&14&140&68&81&3\\
    Center 10&44&68&29&26&21&44&17&151.5&61&90&3\\
    Center 11&37&61&22&37&22&39&16&188.5&70&80&3.5\\
    Center 12&52&75&18&45&11&39&16&184&66&91&3\\
    Center 13&47&69&16&41&16&37&19&121&69&87&2.5\\
    Center 14&40&80&20&30&17&43&15&173.5&63&80&4\\
    Center 15&43&65&39&39&9&30&15&167&74&83&4\\
    Center 16&60&60&10&30&30&30&13&191.5&70&70&3\\
    \hline
    \multicolumn{12}{p{0.95\linewidth}}{Asterisks indicate significant dependence between each variable and hospital centers ($^*$: p $<$ 0.05; $^**$: p $<$ 0.01; $^***$: p $<$ 0.001).}\\
    \multicolumn{12}{p{0.95\linewidth}}{$^\#$ \% of males is indicated.}\\
    \multicolumn{12}{p{0.95\linewidth}}{Abbreviations: HTN: hypertension; DM: diabetes mellitus; AF: atrial fibrillation; SMK: smoking; HCL: hypercholesterolemia; NIHSS: NIH Stroke Scale upon admission; P2P: time interval in minutes between MRI and the arterial puncture; IVT: previous intravenous thrombolysis; RECA: successful recanalisation; mRS\_m3: Modified Rankin Scale at three months after stroke.}\\
    \end{tabular}
    \label{tab:cliByCenter}
\end{table*}

\section{Odds ratios and Wald test in logistic regression of functional outcome}
\label{appendix_oddsRatio}
\setcounter{table}{0}
The contributions of each predictor in logistic regression are summarized in Table \ref{tab:std_OR}.

\begin{table*}
    \footnotesize
    \centering
    \caption{\textbf{Standardized odds ratio in multiple logistic regression of good functional outcome in the multicenter evaluation cohort.}}
    \renewcommand{\arraystretch}{1.6}
    \begin{tabular}{>{\raggedright}p{0.07cm}>{\raggedright}p{0.10\linewidth}>{\raggedright}p{0.03\linewidth}>{\raggedright}p{0.03\linewidth}>{\raggedright}p{0.04\linewidth}>{\raggedright}p{0.03\linewidth}>{\raggedright}p{0.03\linewidth}>{\raggedright}p{0.04\linewidth}>{\raggedright}p{0.04\linewidth}>{\raggedright}p{0.05\linewidth}>{\raggedright}p{0.04\linewidth}>{\raggedright}p{0.04\linewidth}>{\raggedright}p{0.05\linewidth}>{\raggedright\arraybackslash}p{0.09\linewidth}}
        \hline
        &&\textbf{age}&\textbf{sex}$^\#$&\textbf{HTN}&\textbf{DM}&\textbf{AF}&\textbf{SMK}&\textbf{HCL}&\textbf{NIHSS}&\textbf{P2P}&\textbf{IVT}&\textbf{RECA}&\textbf{BrainAGE}\\
        \hdashline
        \multirow{3}{*}{\raisebox{-1.1\height}{\rotatebox[origin=c]{90}{\textbf{VolSimpleLR}}}}&\textbf{centralized}&\specialcell[c]{0.535\\$^{***}$}&\specialcell[c]{1.189\\$^*$}&\specialcell[c]{1.023\\$\quad$}&\specialcell[c]{0.885\\$\quad$}&\specialcell[c]{1.034\\$\quad$}&\specialcell[c]{0.907\\$\quad$}&\specialcell[c]{1.064\\$\quad$}&\specialcell[c]{0.535\\$^{***}$}&\specialcell[c]{0.809\\$^{**}$}&\specialcell[c]{1.462\\$^{***}$}&\specialcell[c]{2.099\\$^{***}$}&\specialcell[c]{0.811\\$^{*}$}\\
        
        &\textbf{federated}&\specialcell[c]{0.536\\$^{***}$}&\specialcell[c]{1.187\\$^*$}&\specialcell[c]{1.024\\$\quad$}&\specialcell[c]{0.880\\$\quad$}&\specialcell[c]{1.028\\$\quad$}&\specialcell[c]{0.907\\$\quad$}&\specialcell[c]{1.067\\$\quad$}&\specialcell[c]{0.536\\$^{***}$}&\specialcell[c]{0.809\\$^{**}$}&\specialcell[c]{1.464\\$^{***}$}&\specialcell[c]{2.105\\$^{***}$}&\specialcell[c]{0.833\\$^*$}\\
        
        &\textbf{single-site}&\specialcell[c]{0.537\\$^{***}$}&\specialcell[c]{1.183\\$^*$}&\specialcell[c]{1.024\\$\quad$}&\specialcell[c]{0.877\\$\quad$}&\specialcell[c]{1.025\\$\quad$}&\specialcell[c]{0.907\\$\quad$}&\specialcell[c]{1.068\\$\quad$}&\specialcell[c]{0.537\\$^{***}$}&\specialcell[c]{0.808\\$^{**}$}&\specialcell[c]{1.465\\$^{***}$}&\specialcell[c]{2.108\\$^{***}$}&\specialcell[c]{0.846\\$^*$}\\

        \hdashline

        \multirow{3}{*}{\raisebox{0\height}{\rotatebox[origin=c]{90}{\textbf{VolAugmentedLR}}}}&\textbf{centralized}&\specialcell[c]{0.536\\$^{***}$}&\specialcell[c]{1.196\\$^*$}&\specialcell[c]{1.034\\$\quad$}&\specialcell[c]{0.884\\$\quad$}&\specialcell[c]{1.029\\$\quad$}&\specialcell[c]{0.909\\$\quad$}&\specialcell[c]{1.062\\$\quad$}&\specialcell[c]{0.534\\$^{***}$}&\specialcell[c]{0.809\\$^{**}$}&\specialcell[c]{1.460\\$^{***}$}&\specialcell[c]{2.098\\$^{***}$}&\specialcell[c]{0.811\\$^{*}$}\\

        &\textbf{federated}&\specialcell[c]{0.535\\$^{***}$}&\specialcell[c]{1.193\\$^*$}&\specialcell[c]{1.033\\$\quad$}&\specialcell[c]{0.882\\$\quad$}&\specialcell[c]{1.026\\$\quad$}&\specialcell[c]{0.905\\$\quad$}&\specialcell[c]{1.068\\$\quad$}&\specialcell[c]{0.537\\$^{***}$}&\specialcell[c]{0.810\\$^{**}$}&\specialcell[c]{1.463\\$^{***}$}&\specialcell[c]{2.103\\$^{***}$}&\specialcell[c]{0.822\\$^*$}\\

        &\textbf{single-site}&\specialcell[c]{0.534\\$^{***}$}&\specialcell[c]{1.196\\$^*$}&\specialcell[c]{1.037\\$\quad$}&\specialcell[c]{0.883\\$\quad$}&\specialcell[c]{1.027\\$\quad$}&\specialcell[c]{0.907\\$\quad$}&\specialcell[c]{1.069\\$\quad$}&\specialcell[c]{0.536\\$^{***}$}&\specialcell[c]{0.809\\$^{**}$}&\specialcell[c]{1.462\\$^{***}$}&\specialcell[c]{2.105\\$^{***}$}&\specialcell[c]{0.806\\$^{*}$}\\

        \hdashline

        \multirow{3}{*}{\raisebox{-1.1\height}{\rotatebox[origin=c]{90}{\textbf{RadiomicsLR}}}}&\textbf{centralized}&\specialcell[c]{0.527\\$^{***}$}&\specialcell[c]{1.127\\$\quad$}&\specialcell[c]{1.025\\$\quad$}&\specialcell[c]{0.895\\$\quad$}&\specialcell[c]{1.030\\$\quad$}&\specialcell[c]{0.900\\$\quad$}&\specialcell[c]{1.080\\$\quad$}&\specialcell[c]{0.530\\$^{***}$}&\specialcell[c]{0.809\\$^{**}$}&\specialcell[c]{1.478\\$^{***}$}&\specialcell[c]{2.119\\$^{***}$}&\specialcell[c]{0.771\\$^{**}$}\\

        &\textbf{federated}&\specialcell[c]{0.526\\$^{***}$}&\specialcell[c]{1.131\\$\quad$}&\specialcell[c]{1.024\\$\quad$}&\specialcell[c]{0.891\\$\quad$}&\specialcell[c]{1.026\\$\quad$}&\specialcell[c]{0.898\\$\quad$}&\specialcell[c]{1.076\\$\quad$}&\specialcell[c]{0.531\\$^{***}$}&\specialcell[c]{0.822\\$^*$}&\specialcell[c]{1.469\\$^{***}$}&\specialcell[c]{2.109\\$^{***}$}&\specialcell[c]{0.774\\$^{**}$}\\

        &\textbf{single-site}&\specialcell[c]{0.522\\$^{***}$}&\specialcell[c]{1.117\\$\quad$}&\specialcell[c]{1.022\\$\quad$}&\specialcell[c]{0.880\\$\quad$}&\specialcell[c]{1.023\\$\quad$}&\specialcell[c]{0.899\\$\quad$}&\specialcell[c]{1.079\\$\quad$}&\specialcell[c]{0.536\\$^{***}$}&\specialcell[c]{0.822\\$^{**}$}&\specialcell[c]{1.477\\$^{***}$}&\specialcell[c]{2.097\\$^{***}$}&\specialcell[c]{0.739\\$^{**}$}\\

        \hdashline

        \multirow{3}{*}{\raisebox{-1.3\height}{\rotatebox[origin=c]{90}{\textbf{VoxelsCNN}}}}&\textbf{centralized}&\specialcell[c]{0.531\\$^{***}$}&\specialcell[c]{1.137\\$\quad$}&\specialcell[c]{1.029\\$\quad$}&\specialcell[c]{0.888\\$\quad$}&\specialcell[c]{1.019\\$\quad$}&\specialcell[c]{0.900\\$\quad$}&\specialcell[c]{1.069\\$\quad$}&\specialcell[c]{0.531\\$^{***}$}&\specialcell[c]{0.807\\$^{**}$}&\specialcell[c]{1.473\\$^{***}$}&\specialcell[c]{2.129\\$^{***}$}&\specialcell[c]{0.783\\$^{**}$}\\

        &\textbf{federated}&\specialcell[c]{0.531\\$^{***}$}&\specialcell[c]{1.137\\$\quad$}&\specialcell[c]{1.019\\$\quad$}&\specialcell[c]{0.879\\$\quad$}&\specialcell[c]{1.034\\$\quad$}&\specialcell[c]{0.901\\$\quad$}&\specialcell[c]{1.076\\$\quad$}&\specialcell[c]{0.540\\$^{***}$}&\specialcell[c]{0.807\\$^{**}$}&\specialcell[c]{1.475\\$^{***}$}&\specialcell[c]{2.107\\$^{***}$}&\specialcell[c]{0.821\\$^{*}$}\\

        &\textbf{single-site}&\specialcell[c]{0.525\\$^{***}$}&\specialcell[c]{1.136\\$\quad$}&\specialcell[c]{1.023\\$\quad$}&\specialcell[c]{0.883\\$\quad$}&\specialcell[c]{1.033\\$\quad$}&\specialcell[c]{0.895\\$\quad$}&\specialcell[c]{1.074\\$\quad$}&\specialcell[c]{0.533\\$^{***}$}&\specialcell[c]{0.807\\$^{**}$}&\specialcell[c]{1.478\\$^{***}$}&\specialcell[c]{2.109\\$^{***}$}&\specialcell[c]{0.764\\$^{**}$}\\
        \hline
        \multicolumn{14}{p{0.95\linewidth}}{$^\#$ Sex was encoded as 1 for males and 0 for females.}\\
        \multicolumn{14}{p{0.95\linewidth}}{Asterisks indicate that the BrainAGE regression coefficient was significantly different from 0 ($^*$: p $<$ 0.05; $^**$: p $<$ 0.01; $^***$: p $<$ 0.001).}\\
        \multicolumn{14}{p{0.95\linewidth}}{Abbreviations: HTN: hypertension; DM: diabetes mellitus; AF: atrial fibrillation; SMK: smoking; HCL: hypercholesterolemia; NIHSS: NIH Stroke Scale upon admission; P2P: time interval in minutes between MRI and the arterial puncture; IVT: previous intravenous thrombolysis; RECA: successful recanalisation.}\\
    \end{tabular}
    \label{tab:std_OR}
\end{table*}

\section{Prediction errors of linear models with FedProx}
\label{appendix_fedprox}
\setcounter{figure}{0}
We investigated the effect of extending FedAvg with the FedProx proximal term \citep{Li2020}, which is designed to improve training under inter-client data heterogeneity. To do so, we retrained our linear models by adding the proximal term with a weight of $\mu = 0.01$. We did not include \textit{VoxelsCNN} due to the high computational cost.

\begin{figure}
\centering
\includegraphics{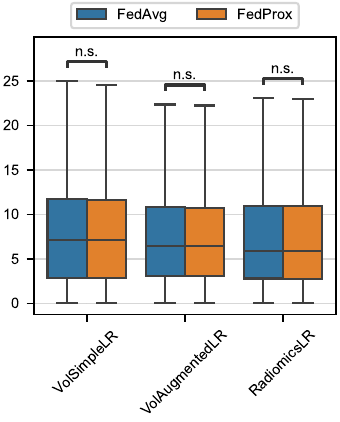}
\caption{\textbf{Absolute age prediction errors in the multicenter evaluation cohort of linear models trained with FedAvg and FedProx.} \textit{n.s.} indicates a non-significant paired comparison of errors ($p > 0.05$).}
\label{fig:fedprox_errs}
\end{figure}

Error distributions were very similar for models trained with FedProx and FedAvg and were never significantly different according to two-tailed Wilcoxon signed-rank test with BH correction (Figure \ref{fig:fedprox_errs}).

\section{BrainAGE and lesion volumes}
\setcounter{figure}{0}
\label{appendix_lesionVolsVSbrainAGE}
Among the 1023 participants in our multicenter evaluation cohort, 134 were manually annotated by radiologists for ischemic stroke lesions, based on diffusion-weighted imaging acquired during the same MRI session as the FLAIR images used in this study. We then examined the relationship between BrainAGE and lesion volume by computing Pearson correlation coefficients and performing the corresponding Student’s t-test with BH correction. The signs of the coefficients were inconsistent across models and none differed significantly from zero (p $>$ 0.05; Figure \ref{fig:lesionVols_brainAGE}).

\begin{figure*}
    \centering
    \includegraphics[scale=1]{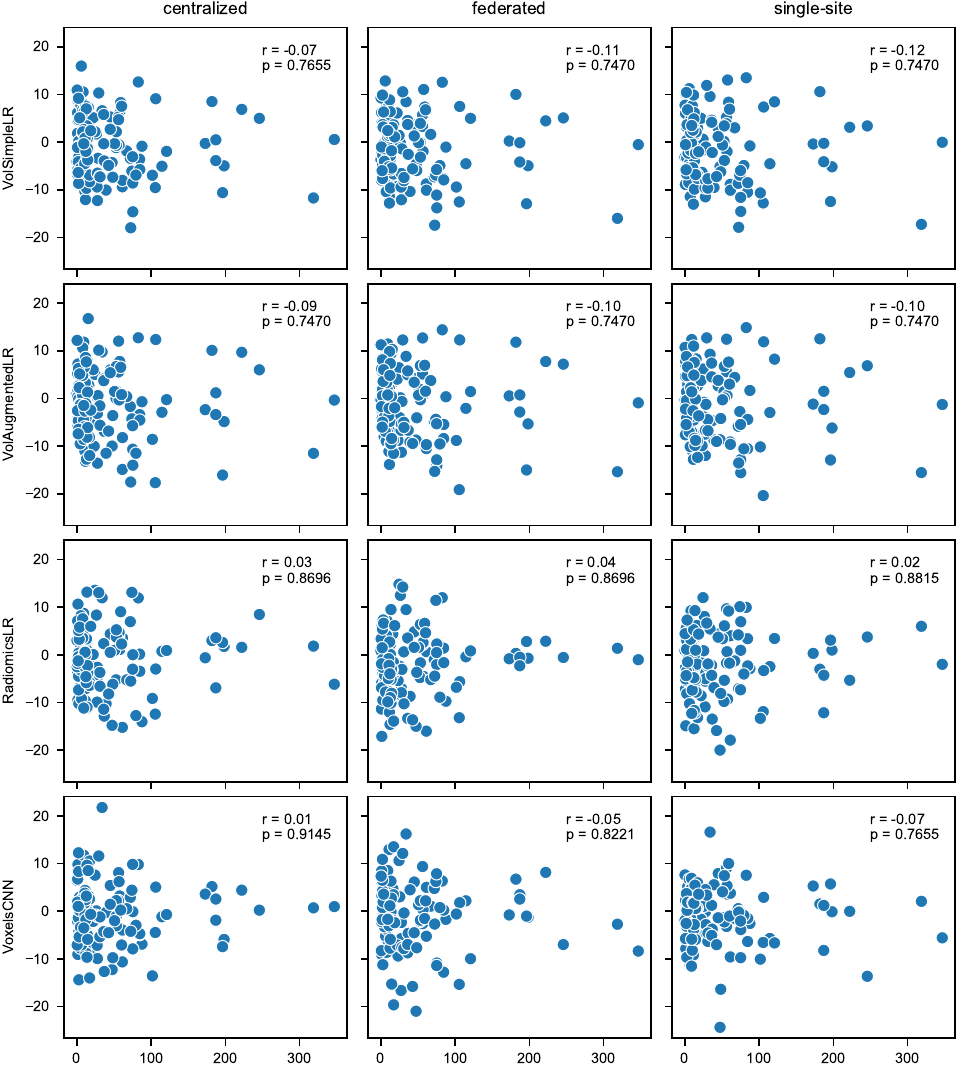}
    \caption{\textbf{BrainAGE as a function of lesion volume in the multicenter evaluation cohort.} 134 patients with manually annotated ischemic stroke lesions are shown. The X- and Y-axes represent lesion size (ml) and BrainAGE, respectively. For each model, the Pearson correlation coefficient (r) and the corresponding p-value (p) are reported.}
    \label{fig:lesionVols_brainAGE}
\end{figure*}

\section{Additional comparisons with a state-of-the-art BrainAGE model}
\label{appendix_sota_comparison}
\setcounter{figure}{0}
To investigate the influence of using clinical training data, we compared our centralized \textit{VoxelsCNN} model with \citet{Colman2026}'s Inception-ResNet-V2\footnote{\url{https://github.com/jordan-colman/FLAIR-Brain-Age/tree/84822e8} accessed 2026-07-17}, which was trained using a dataset of 1568 brain FLAIR images from healthy participants. We applied the model to the multicenter evaluation cohort to obtain age predictions and computed BrainAGE using the same procedure as for our models (Section \ref{meth:brainAge_computation}).  

\begin{figure}
    \centering
    \includegraphics{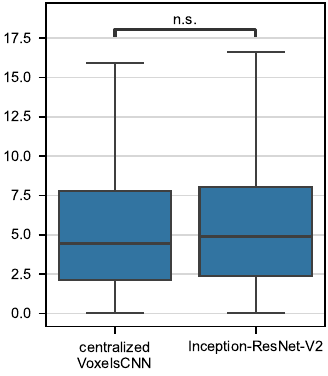}
    \caption{\textbf{Absolute age prediction errors in the multicenter evaluation cohort for the centralized \textit{VoxelsCNN} model and Inception-ResNet-V2.} \textit{n.s.} indicates a non-significant paired comparison of errors ($p \ge 0.05$).}
    \label{fig:colman_errs}
\end{figure}

\begin{figure}
    \centering
    \includegraphics{fig5_legend.pdf}
    \includegraphics{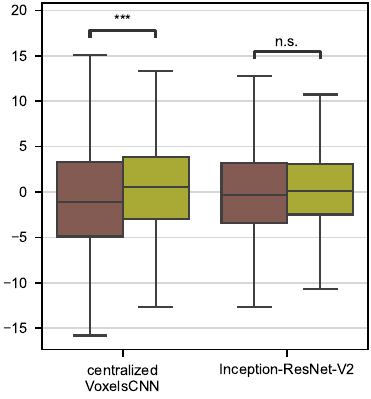}
    \caption{\textbf{Comparison of BrainAGE between good and poor functional outcome in the multicenter evaluation cohort for the centralized VoxelsCNN model and Inception-ResNet-V2.} Asterisks indicate significant unpaired comparisons of BrainAGE ($^*$: p $<$ 0.05; $^{**}$: p $<$ 0.01; $^{***}$: p $<$ 0.001; \textit{n.s.}: p $\ge$ 0.05).}
    \label{fig:colman_brainage}
\end{figure}

We conducted a two-tailed Wilcoxon signed-rank test to compare prediction errors but found no significant differences between the two models (Figure \ref{fig:colman_errs}). However, when comparing BrainAGE between patients with good and poor functional outcome using Mann-Whitney U tests, the centralized \textit{VoxelsCNN} model showed a highly significant difference, whereas the difference was not significant for Inception-ResNet-V2 (Figure \ref{fig:colman_brainage}).

\begin{figure}
    \centering
    \includegraphics{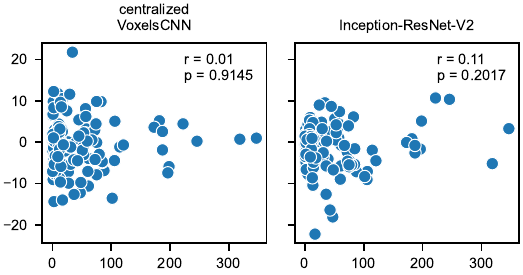}
    \caption{\textbf{BrainAGE as a function of lesion volume in the multicenter evaluation cohort for the centralized VoxelsCNN model and Inception-ResNet-V2.} 134 patients with manually annotated ischemic stroke lesions are shown. The X- and Y-axes represent lesion size (ml) and BrainAGE, respectively. For each model, the Pearson correlation coefficient (r) and the corresponding p-value (p) are reported.}
    \label{fig:colman_brainageVSlesionVol}
\end{figure}

To investigate the cause of this difference, we repeated the analyses performed for our models to quantify the correlation between lesion volume and BrainAGE (\ref{appendix_lesionVolsVSbrainAGE}). Although neither model exhibited a statistically significant correlation, the correlation was stronger for Inception-ResNet-V2 (Figure \ref{fig:colman_brainageVSlesionVol}).

\section{Age distribution per clinical phenotype}
\setcounter{figure}{0}
\label{appendix_agePhenotype}
Figure \ref{fig:age_phenotype} presents age distributions for each clinical phenotype in the multicenter evaluation cohort. Statistical significance of age differences was assessed using two-tailed Mann–Whitney U tests with BH correction.

\begin{figure}
    \centering
    \includegraphics[scale=1]{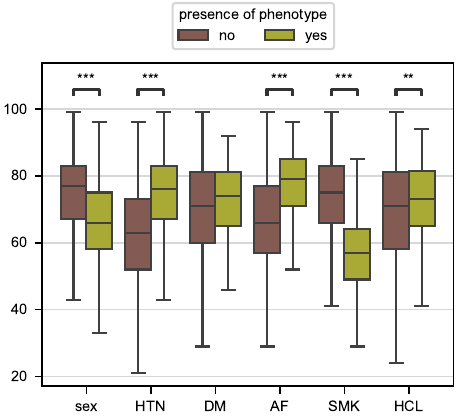}
    \caption{\textbf{Comparison of ages for each clinical phenotype in the multicenter evaluation cohort.} For the sex variable, the presence or absence of the phenotype corresponds to male and female, respectively. Asterisks indicate significant unpaired comparisons of ages ($^*$: p $<$ 0.05; $^**$: p $<$ 0.01; $^***$: p $<$ 0.001).}
    \label{fig:age_phenotype}
\end{figure}

\section{Age-matched BrainAGE comparison for each clinical phenotype}
\label{appendix_ageMatchComparisons}
\setcounter{figure}{0}
We reproduced the experiments described in Section \ref{meth:clinicalPhenotypes} using age-matched comparisons. For each clinical phenotype (sex, HTN, DM, AF, SMK, and HCL), we first separated individuals with and without the phenotype (or male and female for sex). We then subsampled the two groups using a greedy nearest-neighbor matching algorithm based on the absolute age difference.

To reduce order-dependent bias and maximize the number of matched pairs, we repeated the nearest-neighbor matching without replacement ten times, each time iterating through the smaller group in a random order and allowing matches within a broader ±8-year tolerance. Among these repetitions, we retained the configuration yielding the largest number of pairs or, in the case of ties, the smallest total sum of absolute age differences.

\begin{figure*}
    \centering
    \includegraphics[scale=1]{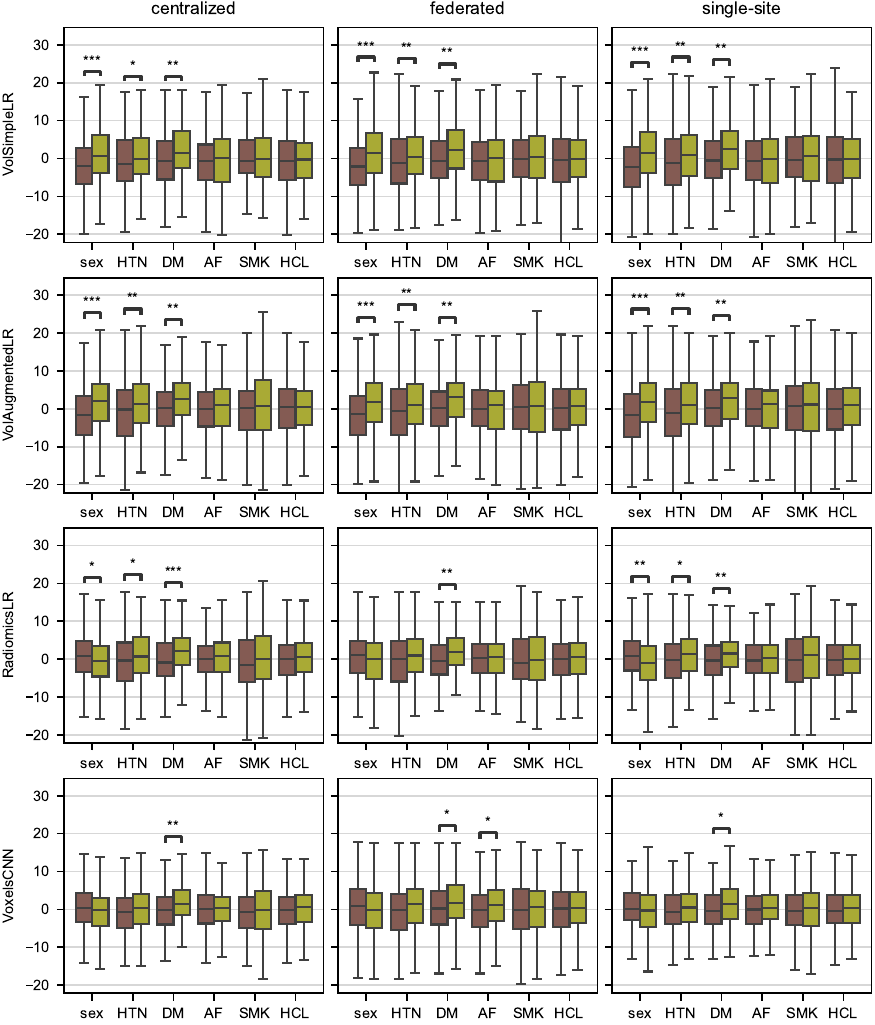}\\
    \includegraphics[scale=1]{fig4_legend.pdf}
    \caption{\textbf{Age-matched comparison of BrainAGE for each clinical phenotype in the multicenter evaluation cohort.} For the sex variable, the presence or absence of the phenotype corresponds to male and female, respectively. Asterisks indicate significant unpaired comparisons (two-tailed Mann-Whitney U tests) of BrainAGE ($^*$: p $<$ 0.05; $^{**}$: p $<$ 0.01; $^{***}$: p $<$ 0.001).\\
    Abbreviations: HTN: hypertension; DM: diabetes mellitus; AF: atrial fibrillation; SMK: smoking; HCL: hypercholesterolemia.\\
    Number of matched pairs: sex: 390; HTN: 288; DM: 217; AF: 344; SMK: 171; HCL: 419.}
    \label{fig:ageMatch_comparisons}
\end{figure*}

The resulting age-matched comparisons of BrainAGE for each clinical phenotype are presented in Figure \ref{fig:ageMatch_comparisons}. Patients with DM exhibited significantly higher BrainAGE across all models. HTN patients also showed higher BrainAGE, with statistical significance in the \textit{VolSimpleLR}, \textit{VolAugmentedLR}, and both centralized and single-site \textit{RadiomicsLR} models. AF patients exhibited significantly higher BrainAGE only in the federated \textit{VoxelsCNN} model. No significant differences were observed for SMK and HCL. Regarding sex, males exhibited significantly higher BrainAGE in the \textit{VolSimpleLR} and \textit{VolAugmentedLR} models. This trend was reversed for the \textit{RadiomicsLR} and \textit{VoxelsCNN} models, reaching statistical significance in the centralized and single-site \textit{RadiomicsLR} models.

\section{Prediction errors across training configurations}
\setcounter{table}{0}
\label{appendix_trainingConfigs}
We conducted additional experiments with different training configurations to examine how data management strategies influence age prediction errors. Due to the substantial computational requirements of these procedures, deep learning models were excluded from these analyses.

\subsection{Leave-one-center-out evaluation for centralized and federated configurations}
To assess the generalization ability of the centralized and federated models, we performed LOCO experiments. In this approach, instead of using the classical cross-validation strategy applied in the main analysis, each model was trained on data from 15 centers and tested on the remaining center in turn.

Under the LOCO setup, prediction errors for the centralized and federated configurations were similar, whereas single-site models still exhibited higher errors (Table \ref{tab:errorsLOCO}).

\begin{table*}
    \centering
    \caption{\textbf{Absolute age prediction errors in the multicenter evaluation cohort using Leave-one-center-out (LOCO) cross-validation for the centralized and federated configurations.}}
    \begin{tabular}{>{\raggedright}p{0.22\linewidth}>{\raggedright}p{0.22\linewidth}>{\raggedright}p{0.22\linewidth}>{\raggedright\arraybackslash}p{0.18\linewidth}}
        \hline
        &\textbf{centralized LOCO}&\textbf{federated LOCO}&\textbf{single-site}\\
        \textbf{VolSimpleLR}&8.14 ± 6.34&8.14 ± 6.42&8.49 ± 6.91\\
        \textbf{VolAugmentedLR}&7.49 ± 5.94&7.56 ± 5.86&7.65 ± 6.07\\
        \textbf{RadiomicsLR}&7.02 ± 5.60&7.29 ± 5.81&8.60 ± 8.65\\
        \hline
        \multicolumn{4}{p{0.9\linewidth}}{Mean ± standard deviation of the prediction errors are reported. The single-site results correspond to those reported in the main study.}
    \end{tabular}
    \label{tab:errorsLOCO}
\end{table*}

\subsection{Within-center evaluation for the single-site configuration}
To evaluate the impact of the different centers between the training and inference phases in the single-site configuration (training on Center 1 and inference on the 15 other centers in the main analysis), we performed an additional experiment in which inference was done separetely for each of the 15 centers in the multicenter evaluation cohort  using within-center cross-validation.

Prediction errors remained higher for the single-site configuration compared with the centralized and federated models (Table \ref{tab:errorsWS}).

\begin{table*}
    \centering
    \caption{\textbf{Absolute age prediction errors in the multicenter evaluation cohort using within-site (WS) cross-validation for the single-site configuration.}}
    \begin{tabular}{>{\raggedright}p{0.22\linewidth}>{\raggedright}p{0.18\linewidth}>{\raggedright}p{0.18\linewidth}>{\raggedright\arraybackslash}p{0.18\linewidth}}
        \hline
        &\textbf{centralized}&\textbf{federated}&\textbf{single-site WS}\\
        \textbf{VolSimpleLR}&8.03 ± 6.22&8.13 ± 6.43&8.71 ± 7.42\\
        \textbf{VolAugmentedLR}&7.44 ± 5.87&7.56 ± 5.85&9.74 ± 8.71\\
        \textbf{RadiomicsLR}&6.85 ± 5.57&7.35 ± 5.97&7.85 ± 6.52\\
        \hline
        \multicolumn{4}{p{0.9\linewidth}}{Mean ± standard deviation of the prediction errors are reported. The centralized and federated results correspond to those reported in the main study.}
    \end{tabular}
    \label{tab:errorsWS}
\end{table*}

\subsection{Different training centers for the single-site configuration}
In the main analysis, Center 1 was used for training in the single-site configuration because it was the most represented center in the dataset. To examine the influence of this choice, we conducted an additional experiment in which models were trained separately on each of the 15 other centers. Using each of the 16 single-site models, we then performed inference on the 15 remaining centers, resulting in 15 age predictions for each participant in the multicenter evaluation cohort. For each participant, the prediction errors were averaged and compared with those obtained from the model trained on Center 1.

Prediction errors were substantially lower when using Center 1 for training compared with the average across all single-site models (Table \ref{tab:errorsAllSingleSite}).

\begin{table*}
    \centering
    \caption{\textbf{Absolute age prediction errors in the multicenter evaluation cohort with different single-site trainings.}}
    \begin{tabular}{>{\raggedright}p{0.28\linewidth}>{\raggedright}p{0.28\linewidth}>{\raggedright\arraybackslash}p{0.28\linewidth}}
        \hline
        &\textbf{single-site}&\textbf{single-site AVG}\\
        \textbf{VolSimpleLR}&8.49 ± 6.91&9.14 ± 7.42\\
        \textbf{VolAugmentedLR}&7.65 ± 6.07&9.79 ± 7.94\\
        \textbf{RadiomicsLR}&8.60 ± 8.65&9.67 ± 17.23\\
        \hline
        \multicolumn{3}{p{0.9\linewidth}}{Mean ± standard deviation of the prediction errors are reported. \textit{single-site} (Center 1) corresponds to models trained on Center 1. \textit{single-site AVG} corresponds to the mean error per participant across the 15 models trained on the other centers.}
    \end{tabular}
    \label{tab:errorsAllSingleSite}
\end{table*}

\bibliographystyle{elsarticle-harv}\biboptions{authoryear}
\bibliography{refs}

\end{document}